\documentclass{article}
\usepackage{spconf,amsmath,graphicx,subfigure}
\usepackage{gensymb}
\usepackage{booktabs}
\usepackage{url}

\title{Visual-Only Recognition of Normal, Whispered and Silent Speech\thanks{Accepted to ICASSP 2018}}
\twoauthors
  { Stavros Petridis, Jie Shen, Doruk Cetin\sthanks{The work was carried out while visiting the iBUG group.} }
{Imperial College London\\
Dept. of Computing, London, UK\\
\{sp104;js1907\}@imperial.ac.uk}
{Maja Pantic }
	{Imperial College London \\
	Dept. of Computing, UK \\
	m.pantic@imperial.ac.uk}

\begin{document}
%
\maketitle
\begin{abstract}
Silent speech interfaces have been recently proposed as a way to enable communication when the acoustic signal is not available. This introduces the need to build visual speech recognition systems for silent and whispered speech. However, almost all the recently proposed systems have been trained on vocalised data only. This is in contrast with evidence in the literature which suggests that lip movements change depending on the speech mode. In this work, we introduce a new audiovisual database which is publicly available and contains normal, whispered and silent speech. To the best of our knowledge, this is the first study which investigates the differences between the three speech modes using the visual modality only. We show that an absolute decrease in classification rate of up to 3.7\% is observed when training and testing on normal and whispered, respectively, and vice versa. An even higher decrease of up to 8.5\% is reported when the models are tested on silent speech. This reveals that there are indeed visual differences between the 3 speech modes and the common assumption that vocalized   training   data can be used directly to train a silent speech recognition system may not be true.

\end{abstract}
\section{Introduction}
\label{sec:intro}

Visual speech recognition is a way of understanding speech by observing only the lip movements without having access to the acoustic signal. Several works have been recently presented \cite{petridis2016deep,end2end_multiview,petridis2017deepVisualSpeech,assael2016lipnet,Chung_2017_CVPR,stafylakis2017combining,Potamianos2003} aiming to recognise visual speech. One application of such a system is in noisy acoustic environments since the visual signal is not affected by noise
and can enhance the performance of speech recognition systems.

Another important application which has been recently proposed is silent speech interfaces (SSI) \cite{denby2010silent}. An SSI
is a system enabling speech communication to take place when 
an  audible  acoustic  signal  is  unavailable. This means that a speaker would be able to mouth words instead of actually uttering them and the SSI would recognise the speech content. This is particularly useful for persons with speaking difficulties or in situations where speaking is not allowed, e.g., during a meeting. 

\looseness - 1
However, in all the previous attempts in visual speech  recognition, all models were trained on videos of normal / vocalised\footnote{The terms normal and vocalised speech are used interchangeably in this study.} speech. Although this might be useful in cases where the speaker really vocalises, e.g., in a noisy environment or when he/she is far away, it is not as useful for SSI.
It is known that lip movements are affected by both the context where speech is produced and the mode of speech. There is evidence that  lip movements tend to increase when speech is produced in noise (Lombard speech) \cite{vsimko2016hyperarticulation,garnier2012effect} and in the case of silent speech \cite{bicevskis2016effects}. The latter has also been confirmed in \cite{janke2010impact} where differences in facial electromyography (EMG) signals were observed between vocalised and silent speech.

In other words, the lip movements in vocalised and silent speech are different and this may degrade the performance of models trained on vocalised speech and tested on silent speech. To the best of our knowledge the only work which has addressed this issue, but on a rather small database, is \cite{florescu2010silent}. They used ultrasound tongue images and video lip images from 4 participants and reported a significant drop in performance when training and testing was performed on normal and silent speech, respectively. Similar conclusions have also been observed for normal and whispered speech, which can be thought of as an alternative to silent speech for SSI. The performance of models trained on normal speech decreases when tested on whispered speech \cite{tao2014lipreading,fan2011audio}.

In this work, we introduce a new audiovisual database which contains normal, whispered and silent speech. We recorded 53 participants from 3 different views (frontal, 45\degree and profile) pronouncing digits and phrases in three speech modes. To the best of our knowledge, this is the first audiovisual database which is publicly available and contains all three speech modes. Tran et al. \cite{tran2013audiovisual} have recorded an audiovisual database of normal and whispered speech but without including silent speech. In addition, their
database contains fewer participants, 40, and it is not publicly available at the moment.

We also investigate the differences between the 3 speech modes. We conduct subject independent experiments using an end-to-end lipreading model where we train on one speech mode and test on all other modes. To the best of our knowledge, this is the first study that systematically investigates the differences between the 3 speech modes from the visual modality. Results on digits and phrases demonstrate that there are indeed differences between the speech modes. An absolute decrease between 3.3\% and 3.7\% in classification rate is observed when we train on normal speech and test on whispered speech, and vice versa. A higher absolute decrease in classification rate between 5.7\% and 8.5\% is reported when training on normal or whispered speech and testing on silent speech. Silent speech is consistently the worst performing mode and even when a model is trained and tested on it the performance is lower than the corresponding matched conditions in other speech modes. This is an indication that realising SSIs using the visual modality only is not as straightforward as previously thought since it seems normal speech data may not be enough for training silent speech recognisers.

\begin{figure}[t]

\begin{minipage}[t]{0.95\linewidth}
  \centering
  \subfigure[S002]{\includegraphics[width=0.4\linewidth]{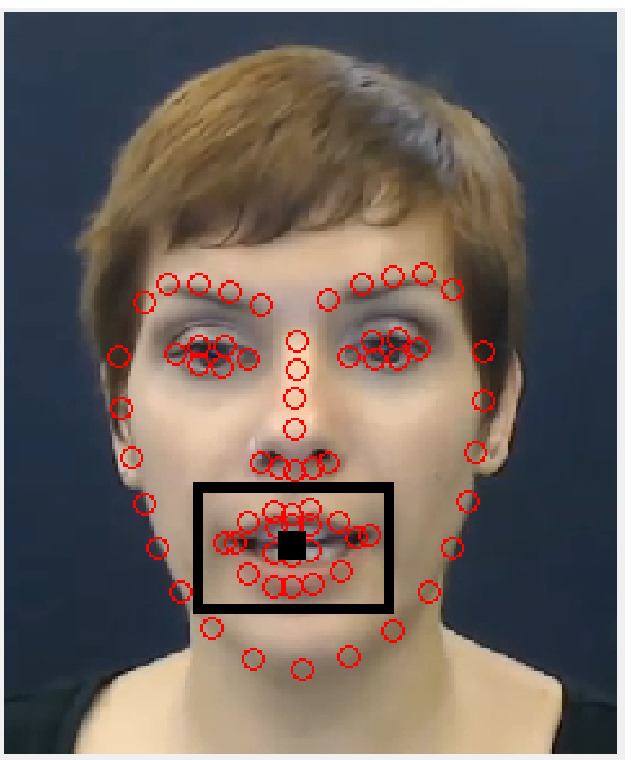}}  
  \subfigure[S012]{\includegraphics[width=0.4\linewidth, trim = 0 0 0 12,clip] {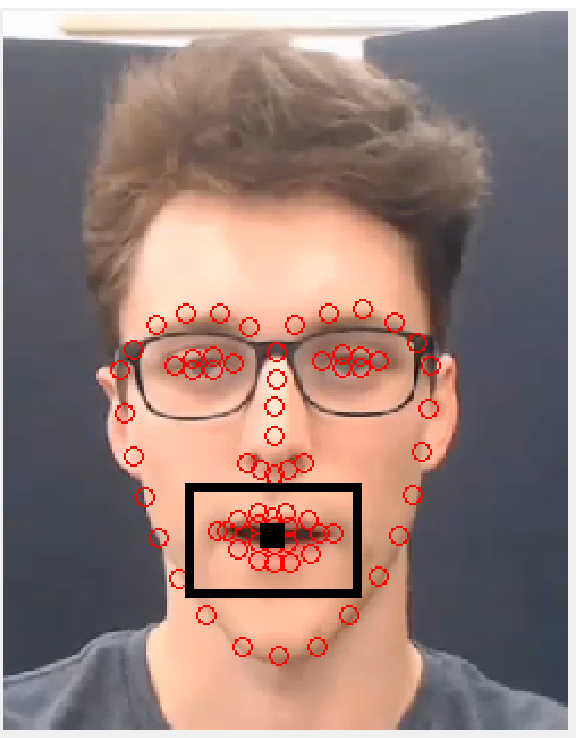}}   

\caption{Example of mouth ROI extraction for participants S002 and S012.}
\label{fig:mouthROI}
\end{minipage}

\end{figure}

\begin{figure}[t]

  \centering
\includegraphics[width=0.6\linewidth]{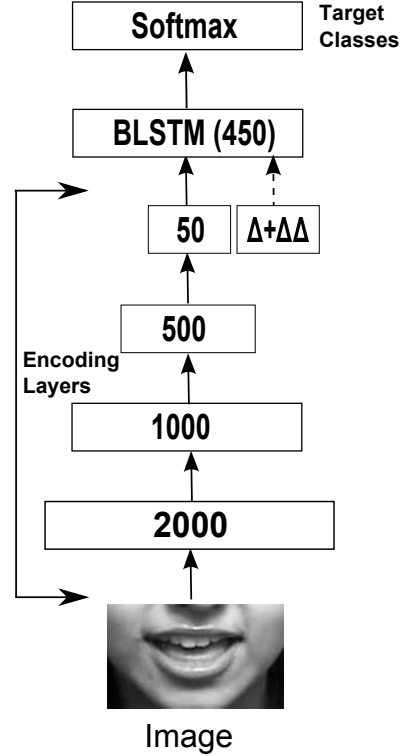}

\caption{Overview of the end-to-end visual speech recognition system. Features are extracted directly from the raw mouth ROI. The $\Delta$  and $\Delta\Delta$  features are also appended to the bottleneck layer. The temporal dynamics are modelled by a BLSTM. }
\label{fig:system}
\end{figure}

\section{Database Description}
\label{sec:database}
For the purposes of this study we have recorded a new audiovisual database which contains normal, whispered and silent speech. The database consists of two parts: digits and short phrases. In the first part, participants were asked to read 10 digits, from 0 to 9, in English in random order five times. They read the digits in three different modes, normal, whispered and silent speech. The only instructions they were given were to speak normally in case of normal speech, to whisper as they would normally whisper and not to produce any audible sound in silent speech. In case of non-native English speakers this part was also repeated in the participant's native language. In total, 53 participants (41 males and 12 females) from 16 nationalities, were recorded with a mean age and standard deviation of 26.7 and 4.3 years, respectively.

In the second part, participants were asked to read 10 short phrases. The phrases are the same as the ones used in the OuluVS2 database \cite{Anina2015}: ``Excuse me'', ``Goodbye'', ``Hello'', ``How are you'', ``Nice to meet you'', ``See you'', ``I am sorry'', ``Thank you'', ``Have a good time'', ``You are welcome''. Again, each phrase was repeated five times in 3 different modes, normal, whisper and silent speech. Thirty nine participants (32 males and 7 females) were recorded for this part with a mean age and standard deviation of 26.3 and 3.8 years, respectively.

The database was recorded in a lab environment using 3 cameras with resolution of 1280 by 780 at 30 frames per second. The 3 cameras record three different views of the participant's face, frontal, 45\degree and profile view, respectively. Audio is also recorded by the 3 cameras using the built-in microphones at 44.1 kHz. 

The digits and phrases were displayed in a laptop screen in front of the participant using slides. Additional information regarding the speech mode to be used was also present in the slide. The participants pressed the space bar after each utterance in order to proceed to the next slide. The space bar hit was used to segment the different digits and phrases. Finally, transcriptions for the digits and phrases were extracted from the slides. The database together with annotations and transcription is already publicly available  \footnote{https://ibug-avs.eu/}. An example of the frontal view recordings from 2 participants in shown in Fig. \ref{fig:mouthROI}.

\section{End-to-end Lipreading}
The deep learning lipreading system used in this study is shown in Fig. \ref{fig:system} and is similar to the one presented in \cite{petridis2017deepVisualSpeech}. The main difference is that we use a single stream instead of two streams since we found no additional performance benefits by the use of the second stream as proposed in \cite{petridis2017deepVisualSpeech}. 

The single stream consists of two parts: an encoder and a BLSTM. The encoder follows a bottleneck architecture in order to compress the high dimensional input image to a low dimensional representation at the bottleneck layer.  The same architecture as in \cite{hinton2006reducing} is used, with 3 hidden layers of sizes 2000, 1000 and 500, respectively, followed by a linear bottleneck layer. The rectified linear unit is used as the activation function for the hidden layers. The $\Delta$ (first derivatives) and $\Delta\Delta$ (second derivatives) \cite{young2002htk} features are also computed, based on the bottleneck features, and they are appended to the bottleneck layer. In this way, during training we force the encoding layers to learn compact representations which are discriminative for the task at hand but also produce discriminative $\Delta$ and $\Delta\Delta$ features. 

The second part is a BLSTM layer added on top of the encoding layers in order to model the temporal dynamics of the features. The output layer is a softmax layer which provides a label for each input frame. The majority label over each utterance is used in order to label the entire utterance. In other words, we follow a classification approach where the models classifies the entire utterance in one out of the ten digits/phrases.

\section{EXPERIMENTAL SETUP}

\subsection{Mouth ROI Extraction}

Sixty eight points are tracked on the face using the tracker proposed in \cite{Kazemi_2014_CVPR}. The faces are first aligned using a normal reference frame in order to normalise them for rotation and size differences. This is done using an affine transform using 6 stable points, the mouth center, two eyes corners in each eye and two points on the nose. Then the center of the mouth is located based on the tracked points and a bounding box
with size 68 by 108 is used to extract the mouth region of interest (ROI) as shown in Fig. \ref{fig:mouthROI}. Finally, the mouth ROIs are downscaled to 32 by 50.

\subsection{Evaluation Protocol}

\looseness - 1
We first partition the data into training, validation and test sets. 
We follow a subject independent scenario where 30, 10, 13 participants are used in the training, validation and test sets, respectively, for the digits experiments. This means that there are 1500 training utterances, 500 validation utterances and 650 test utterances. For the second set of experiments, i.e., phrases, we use 20, 8, 11 participants in training, validation and test sets, respectively. Hence, there are 1000 training utterances, 400 validation utterances and 550 test utterances. The participants used in each set can be found on the database website (see Section \ref{sec:database}). For all the experiments in this study, we used the digits (in English) and phrases in all 3 modes from the frontal camera only.

\subsection{Preprocessing}

Since all the experiments are subject independent we first need to reduce the impact of subject dependent characteristics. This is done by subtracting the mean image, computed over the entire utterance, from each frame.

\looseness - 1
The next step is the normalisation of data.  Each image is z-normalised, i.e. the mean and standard deviation should be equal to 0 and 1 respectively, before training an RBM with linear input units \cite{hinton2012practical}. 

Finally, due to randomness in initialisation, every time a deep network is trained the results are slightly different. In order to present a more objective evaluation we run each experiment 10 times and we report the mean and standard deviation of the classification rate. 

\looseness - 1
 
\subsection{Training}

\looseness - 1
\textbf{Initialisation:} The encoding layers are pre-trained using Restricted Boltzmann Machines (RBMs) \cite{hinton2012practical}.  Since the input (pixels) is real-valued and the hidden layers are either rectified linear or linear (bottleneck layer) four Gaussian RBMs \cite{hinton2012practical}  are used. Each RBM is trained for 20 epochs with a mini-batch size of 100, learning rate of 0.001 and L2 regularisation coefficient of 0.0002 using contrastive divergence. 

\noindent
\textbf{End-to-End Training:} Once the encoder has been pretrained then the BLSTM is added on top and its weights are initialised using glorot initialisation \cite{glorot2010understanding}.
The Adam training algorithm \cite{kingma2014adam} is used for end-to-end training with a mini-batch size of 10 utterances and a learning rate of 0.0003. Early stopping with a delay of 5 epochs was also used in order to avoid overfitting.

\section{Results}

\begin{table}[tb]
\renewcommand{\arraystretch}{1.1}
\caption{Mean classification rate (and standard deviation) for the digits experiment. Each row (column) corresponds to the data the system was trained (tested) on.}
\label{tab:resultsDigits}
\centering
\begin{tabular}{cccc}
\toprule Tested on $\rightarrow$ & Normal  & Whispered & Silent  \\
Trained on $\downarrow$  \\
\midrule Normal & 68.0 (2.1) & 64.7 (1.3) & 59.7 (1.0) \\
Whisper & 66.9 (1.7) & 70.5 (1.3) & 62.8 (2.0) \\
Silent &57.4 (1.6) & 62.1 (1.4) & 62.2 (0.9)\\
\bottomrule

\end{tabular} 

\end{table}

\begin{table}[tb]
\renewcommand{\arraystretch}{1.1}
\caption{Mean classification rate (and standard deviation) for the phrases experiment. Each row (column) corresponds to the data the system was trained (tested) on.}
\label{tab:resultsPhrases}
\centering
\begin{tabular}{cccc}
\toprule Tested on $\rightarrow$ & Normal  & Whispered & Silent  \\
Trained on $\downarrow$  \\
\midrule Normal & 69.7 (2.1) & 66.3 (2.6) & 61.2 (1.6) \\
Whisper & 67.1 (2.8) & 70.8 (2.2) & 65.1 (1.5) \\
Silent &61.0 (1.8) & 65.2 (0.9) & 64.4 (2.3)\\
\bottomrule

\end{tabular} 

\end{table}

Table \ref{tab:resultsDigits} shows the results for the digits experiments. As expected training and testing on the same speech mode leads to the best performance in all 3 cases. On the other hand, training and testing on mismatched modes results in degraded performance. For example, the performance of a model trained on normal speech drops by 3.3\% and 8.3\% when tested on whispered and silent speech examples, respectively. A similar drop is observed when a model trained on whispered speech is tested on normal and silent speech examples. This in line with previous results where the performance of a visual speech recogniser degrades when tested on whispered speech \cite{tao2014lipreading,fan2011audio} and on silent speech \cite{florescu2010silent}.
It is also interesting to point out that performance on silent speech is always low even in the matched condition. 
 
 \looseness -1
 Results for the phrases experiments are shown in Table \ref{tab:resultsPhrases}. It is obvious that all the results are slightly higher than the ones in Table \ref{tab:resultsDigits}. This is possibly due to the longer duration of the phrases, i.e., more information is available. We should also note that the results obtained are significantly lower than the ones obtained on OuluVS2, 91.8\% in \cite{end2end_multiview}, using the same utterances. This is due to the fully automatic extraction of the mouth ROIs and the presence of some segmentation errors \footnote{There are some cases where the participants hit the space bar while uttering a sentence or waited for too long before proceeding to the next slide. We are in the process of manually correcting all the timestamps.}. On the other hand, perfect segmentation and perfectly cropped mouth ROIs (by manually correcting the landmarks if they were off the desired location) are provided with the OuluVS2 database.
 
 Similar conclusions to the digits experiments can be drawn. The performance of a model trained on normal (whispered) speech drops by 3.4\% (3.7\%) when tested on whispered (normal) speech.
 Training on normal speech and testing on silent speech results in a 8.5\% absolute decrease in the classification rate. Similarly, training on whispered speech and testing on silent speech results in a 5.7\% absolute decrease. Again, the performance on silent speech is consistently lower no matter which data the model was trained on. This is probably due to the lack of auditory feedback during articulation which is crucial. As shown in \cite{janke2010impact} using EMG signals the lack of acoustic feedback in silent speech is compensated by a stronger focus on somatosensoric feedback. This is achieved by articulating stronger those sounds which provide more tactile feedback. 
 
Overall, our results agree with the observations in the phonetics literature \cite{bicevskis2016effects} that lip movements are different in the three speech modes. This needs to be taken into account when training visual speech recognisers and it might have a significant impact on silent speech interfaces. Our results suggest that
using a vocalized training data for training a silent speech recognition system, which is very common, results in a significant performance drop.

\section{Conclusion}

In this work, we introduce a new audiovisual database for normal, whispered and silent speech. Results on subject independent experiments reveal that there are differences in lip movements, as it has already been reported in the phonetics literature, which affect the performance of models when the training and testing speech modes are different. In particular, the performance on silent speech suffers the most which indicates that the common approach of using normal visual speech in order to train silent speech recognisers is not the best. Finally, it would be interesting to explore the performance of an audiovisual system on normal and whispered speech and use the side views available in the database in an effort the improve the performance on silent speech.

\section{Acknowledgements}
This work has been funded by the European Community Horizon 2020 under grant agreement
no. 645094 (SEWA).


%
%
%




\bibliographystyle{IEEEbib}
\bibliography{refs}

\end{document}